\documentclass[runningheads]{llncs}

\usepackage{eccv}

\usepackage{eccvabbrv}

\usepackage{graphicx}
\usepackage{booktabs}
\usepackage{multirow}

\usepackage[accsupp]{axessibility}  

\usepackage{hyperref}

\usepackage{orcidlink}

\begin{document}

\title{SCAR-GS: Spatial Context Attention for Residuals in Progressive Gaussian Splatting} 

\titlerunning{SCAR-GS}

\author{Diego Revilla\inst{1, 2}\orcidlink{0009-0005-9004-940X} \thanks{Corresponding author.} \and
Pooja Suresh\inst{1}\orcidlink{0009-0003-2112-5670} \and
Ooi Wei Tsang\inst{1}\orcidlink{0000-0001-8994-1736} \and
Anand Bhojan\inst{1}\orcidlink{0000-0001-8105-1739}
}

\authorrunning{D.Revilla et al.}

\institute{National University of Singapore, 21 Lower Kent Ridge Road, Singapore 119077 \and
University of Deusto,  Avda. de las Universidades 24, 48007 Bilbao, Spain
}

\maketitle

\begin{abstract}
Recent advances in 3D Gaussian Splatting (3DGS) have enabled real-time, high-fidelity novel view synthesis. However, these models incur significant storage requirements for medium and large-scale scenes, limiting their deployment in cloud and streaming applications for Virtual Reality (VR), spatial computing, and real-time 3D rendering on commodity hardware. To overcome the transmission and rendering bottlenecks of large-scale 3D scenes, progressive strategies first transmit coarse scene representations, enabling early rendering and subsequent refinement of either the entire scene or view-dependent regions. This introduces a temporal dimension to spatial scene delivery, where fidelity improves as refinement layers are received, allowing clients to trade latency-to-first-render against final quality under variable bandwidth. Recent compression works make use of scalar quantization to reduce the storage size, without compromising much of the resulting rendering quality. While effective, that presents two key limitations: (1) it cannot exploit similarity between features to avoid redundant transmission, and (2) it fails to leverage reduced-variance representations in learned features, potentially limiting rate-distortion performance.
\\\\
In this work, we introduce a novel progressive codec for 3DGS that replaces traditional methods with Residual Vector Quantization (RVQ) \cite{zeghidour2021soundstream} to compress latent anchor features. This approach decomposes a scene into a coarse base structure and successive high-frequency refinements. Our key contribution is a multi-resolution hash-grid queried autoregressive entropy model that predicts the probability distribution of each transmitted index conditioned on the previously transmitted sequence. Across MipNeRF360 and BungeeNeRF, SCAR-GS comes close to state-of-the-art perceptual quality while offering smoother, lower overhead refinements that scales better to medium and large scenes,  proving well suited to bandwidth-adaptive XR and cloud-streaming.
\end{abstract}
\section{Introduction}
\label{sec:intro}

\begin{figure}
    \centering
    \includegraphics[width=\textwidth]{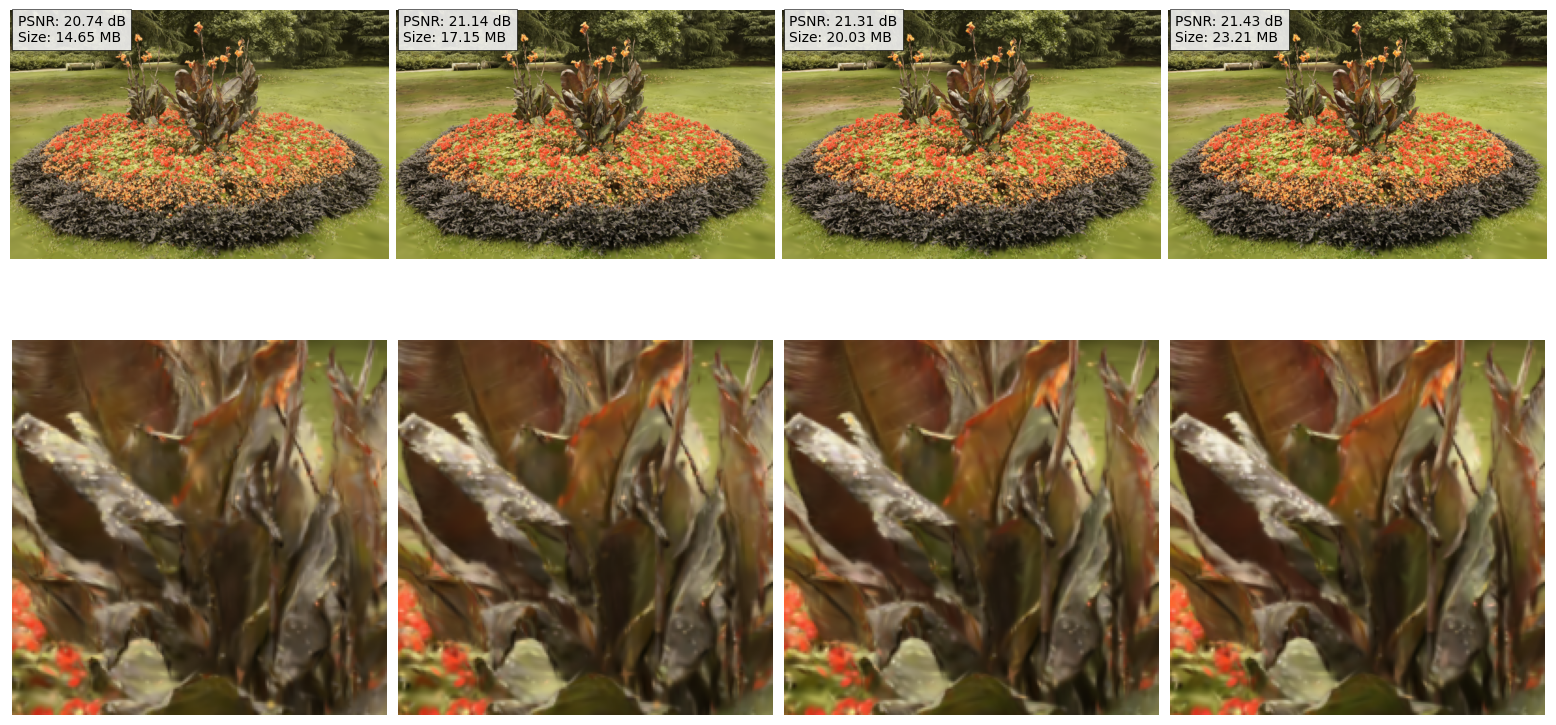}
    \caption{Comparison of progressive refinement layers on the Flower scene from the Mip-NeRF360 dataset \cite{mipnerf360}, showcasing the different RVQ quality refinement for each layer.}
\end{figure}

Gaussian Splatting \cite{kerbl20233dgaussiansplattingrealtime} represents a significant advancement in real-time computer graphics and scene reconstruction, enabling real-time photorealistic rendering and novel-view synthesis beyond traditional Neural Radiance Fields (NeRFs) \cite{10.1007/978-3-030-58452-8_24}. By representing scenes as a collection of scattered Gaussians, this approach offers a  discretized alternative approach to deep neural network inference. However, this explicit representation comes at a cost: Storing Gaussian attributes can be substantial, often reaching hundreds of megabytes per scene \cite{3DGSzip2025}. Such large memory requirements pose a significant barrier to deployment on resource-constrained platforms, including mobile devices and web browsers.

To address this challenge, various compression techniques have been proposed. Most of these methods either prioritize maximum compression efficiency at the expense of granular bitrate control or emphasize progressiveness at the cost of reconstruction quality and coding efficiency. Only a few works \cite{pcgs2025, Sario2025GoDeGO} explore progressive encoding to prevent bottlenecks during memory or network transmission. This is particularly important when latency-to-first-render matters more than early image quality or when handling very large scenes, where distant objects contribute little to perceived rendering quality.

While context modeling has demonstrated remarkable results in compressing 3D Gaussian Scenes, these approaches primarily produce single-rate representations unsuitable for progressive streaming. For instance, HAC \cite{hac2024} and its successor HAC++ \cite{hac++2025} leverage  neural anchors from Scaffold-GS \cite{scaffoldgs} to learn sparse hash grids that capture spatial context. Similarly, ContextGS \cite{10.5555/3737916.3739547} uses an autoregressive model to reuse decoded anchors for predicting finer details. Nevertheless, these methods mainly exploit spatial redundancy in a static reconstruction and do not provide a hierarchical quality representation necessary for progressive transmission. 

Conversely, methods explicitly designed for progressivity often compromise reconstruction quality or compression efficiency. LapisGS \cite{shi2024lapisgs} constructs a layered structure of cumulative Gaussians to incrementally increase rendering resolution, and GoDe \cite{Sario2025GoDeGO} organizes primitives into hierarchical layers based on visibility heuristics. A key limitation of these approaches is additive primitive layering, which transmits new geometric attributes for each refinement stage, incurring structural redundancy. Progressive Compression of Gaussian Splatting (PCGS) \cite{pcgs2025}, introduces a progressive encoding framework using trit-plane encoding and masking to transmit Gaussian attributes in order of importance, allowing control over quantity and quality across layers. However, because it relies on scalar quantization, it may not fully exploit inter-dimensional correlations in feature vectors.

 In this work, we propose a novel progressive codec that replaces independent scalar quantization with a more powerful RVQ \cite{7552944} scheme. 
\\\\
Our contributions are as follows:
\begin{enumerate}
    \item We propose \textbf{SCAR-GS}: \textbf{S}patial \textbf{C}ontext \textbf{A}ttention for \textbf{R}esiduals, a progressive 3DGS codec that conditions on a residual hierarchy to minimize the entropy of transmitted residuals.
    \item We introduce hierarchical vector quantization for 3DGS, including a dual-codebook weight sharing strategy to reduce RVQ-VAE parameter overhead.
    \item We demonstrate through extensive evaluation on standard dataset benchmarks that our RVQ-VAE-based approach achieves reasonable rate–distortion trade-offs in SSIM \cite{1284395} and LPIPS \cite{8578166}, while providing structured progressive refinement with moderate storage growth
    \item We enable progressive spatial scene delivery, where users receive a usable coarse spatial representation first, and successive refinement layers improve perceptual quality over transmission time in chunks.
\end{enumerate}

While PCGS \cite{pcgs2025} achieves a high peak signal-to-noise ratio (PSNR) \cite{gonzalez2002digital} on small scenes due to low overhead, SCAR-GS is designed to provide better scaling for medium-to-large scenes. By utilizing latent RVQ \cite{zeghidour2021soundstream}, we achieve smaller model sizes for complex, unbounded scenes (e.g., Mip-NeRF 360) and enable structured refinement via latent residuals, offering a complementary approach to scalar-based methods.

\section{Related Work}

\subsection{Gaussian Splatting}
3DGS represents a scene using a myriad of 3D Gaussians, each with its own shape and color attributes. Unlike NeRFs, 3DGS enables high-speed rasterization as it does not require real-time network inference. However, this comes at the cost of a significant memory footprint, often reaching hundreds of megabytes per scene.

Initial compression methods such as LightGaussian \cite{lightgaussian} employ significance metrics to remove Gaussians that minimally contribute to the final image. Likewise, compaction methods such as GaussianSpa \cite{zhang2025gaussianspa} try to sparsify the Gaussian scenes in order to reduce unnecessary duplicates. However, both approaches do have an impact on the final image quality. To mitigate this, Scaffold-GS \cite{scaffoldgs} introduces an anchor-based method that not only reduces the final compressed size but also improves rendering quality compared to the original 3DGS baseline. HAC \cite{hac2024} and HAC++ \cite{hac++2025} extend this approach by introducing entropy-focused techniques to further compress the anchor-based representation. Despite these advances, most 3DGS compression research focuses on single-rate representations rather than progressive, layered approaches.

\subsection{Encoded Representations}
Encoded representations allow information to be represented in a compact latent format from which the original data can be reconstructed. Deep representation learning has evolved from continuous latent variable models, such as Variational Autoencoders (VAEs), to discrete variants like Vector-Quantized VAEs (VQ-VAEs) \cite{rvqvae}. By mapping inputs to a finite codebook of learnable embeddings, VQ-VAEs enable efficient latent-space storage.

To address the limited expressivity of a single discretized pass, RVQ \cite{zeghidour2021soundstream} recursively quantizes residual errors across multiple stages, decomposing a signal into a coarse base and a series of high-frequency refinements. Finally, RVQ-VAEs \cite{9879532} extend this paradigm to latent space representations, allowing one or several Gaussians to be represented in a dimensionality-reduced form, which can then be regressed back into fully reconstructed attributes.

In Gaussian Splatting Compression, Scaffold-GS \cite{scaffoldgs} was the first to introduce anchor-level representation learning for nearby Gaussians. Building on this, ContextGS \cite{10.5555/3737916.3739547} predicts higher-order, fine-detail anchors from a coarse set, enabling more accurate and compact representations.

\subsection{Autoregressive Entropy Modeling}
In 3DGS compression, context modeling has proven effective for reducing redundancy \cite{scaffoldgs, 10.5555/3737916.3739547, hac2024, hac++2025}. However, most existing context models are designed for static, single-rate decoding and do not account for the hierarchical requirements of progressive streaming, where context evolves as new refinement layers are received. PCGS \cite{pcgs2025} is the first approach to introduce an autoregressive conditional entropy model for progressive 3DGS compression. Using trit-plane quantization for each encoded dimension, it predicts details in subsequent layers. However, PCGS treats each feature value independently and does not exploit intra-feature correlations, potentially limiting compression efficiency.

\begin{figure}[t]
    \centering
    \includegraphics[width=1\linewidth]{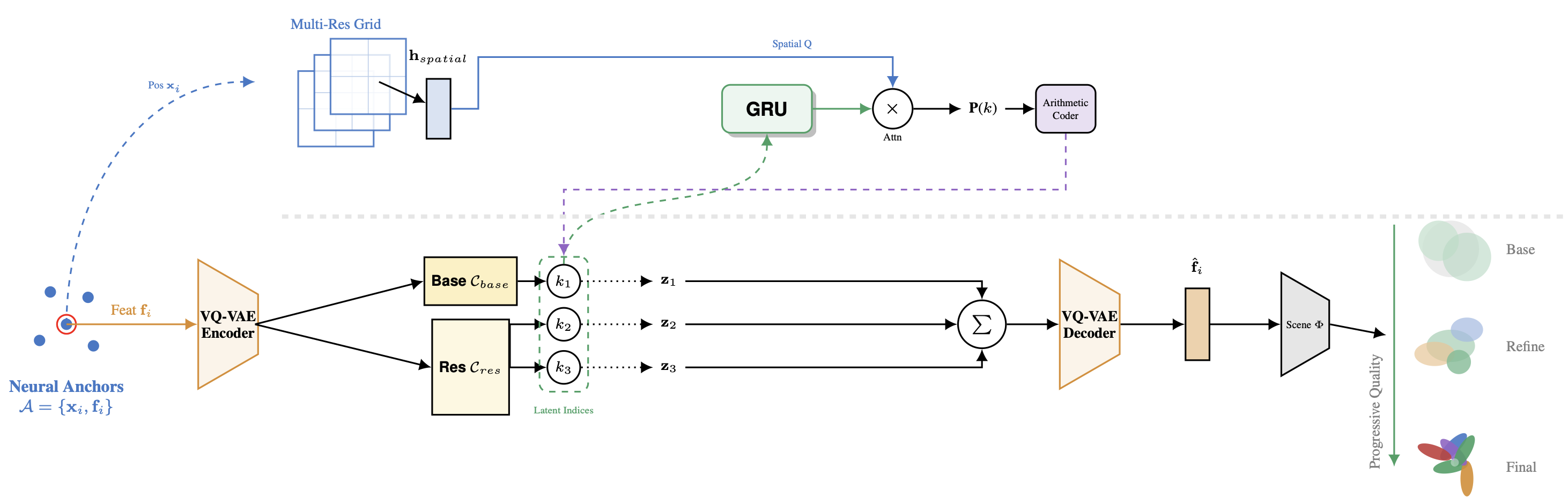}
\caption{Visual overview of the SCAR-GS pipeline. A hierarchical neural representation is encoded using dual-codebook RVQ(Bottom), while a spatially aware autoregressive entropy model(Top) predicts indices for arithmetic coding. The decoder progressively reconstructs features to render Gaussians with increasing fidelity.}
\label{fig:pipeline}
\end{figure}

\section{Methodology}
As depicted in Figure \ref{fig:pipeline}, our approach employs an autoregressive entropy model that operates on the sequence of RVQ indices. Guided by a spatial hash-grid context and previously decoded features, the model predicts the conditional probability of each successive codebook index. This enables a scene representation consisting of a coarse base layer followed by increasingly detailed refinement layers.

\subsection{Preliminaries}
Based on previous works \cite{scaffoldgs, hac2024, hac++2025, pcgs2025}, we represent the volumetric scene $\mathcal{S}$ as a sparse point cloud of $N$ neural anchors clustering nearby Gaussians: 
$\mathcal{A} = \{ \mathbf{x}_i, \mathbf{o}_i, \mathbf{s}_i, \mathbf{f}_i \}_{i=1}^N$, 
where $\mathbf{x}_i \in \mathbb{R}^3$ denotes the anchor position, $\mathbf{o}_i$ the positional offset, $\mathbf{s}_i$ the scaling factor, and $\mathbf{f}_i \in \mathbb{R}^D$ a latent feature encoding local appearance and geometry.

\subsection{Residual Vector Quantization}
We propose to quantize the latent features in the anchors $\mathbf{f}_i$ using an RVQ \cite{zeghidour2021soundstream}. RVQ decomposes feature complexity into a sequence of progressively lower-entropy distributions, making each stage more amenable to accurate conditional probability estimation than a single large codebook. Concretely, RVQ progressively minimizes the reconstruction error across $M$ quantization stages using multiple codebooks.

Standard RVQ uses a single or per-layer codebook, but the initial stage captures high-variance signals, while subsequent stages encode similar residuals. To reduce parameters while maintaining fidelity, we adopt a dual-codebook strategy: a Coarse Codebook $\mathcal{C}_{coarse}$ and a Shared Residual Codebook $\mathcal{C}_{residual}$.

\subsection{The Rotation Trick for gradient propagation}

Since VQ is non-differentiable, we typically rely on the Straight-Through Estimator (STE) where gradients bypass the discretization layer. However, this approach discards critical information about the reconstructed feature locality with respect to the original, potentially leading to poor semantic representation after quantization. To address this, we made use of the Rotation Trick \cite{fifty2025rotationtrick} for gradient propagation. 

\subsection{Spatial-Query Attention Mechanism}
Dependencies across quantization stages are modeled via a multi-layer GRU predicting the next codebook index $k_m$ conditioned on history $k_{<m}$. To incorporate spatial context, we use Spatial-Query Attention: the static spatial embedding is the Query ($Q$), and GRU hidden states form Keys ($K$) and Values ($V$): The attention context $\mathbf{c}_{attn}$ is computed as:$$Q = W_Q \mathbf{h}_{spatial}, \quad K = W_K \mathbf{G}_{<m}, \quad V = W_V \mathbf{G}_{<m}$$$$\mathbf{c}_{attn} = \text{Softmax}\left(\frac{Q K^\top}{\sqrt{d_{model}}}\right) V$$where $\mathbf{G}_{<m}$ represents the sequence of GRU hidden states corresponding to the previous indices, ${h}_{spatial}$ the spatial hash-grid embedding, and $d_{model}$ the dimensionality of the latent space.

\subsection{Optimization Objective}
\subsubsection{Rendering Loss}
The main objective of any 3D Gaussian training framework is to minimize the rendering error between the rendered image and the ground truth. 
$$\mathcal{L}_{scene} = (1 - \lambda_{ssim}) \mathcal{L}_1(I_{rend}, I_{gt}) + \lambda_{ssim} (1 - \text{SSIM}(I_{rend}, I_{gt}))$$ where $\lambda_{ssim}$ is a weighting hyperparameter.

\subsubsection{Entropy Loss}
Given the sequence of ground-truth quantization indices \\$\mathbf{k} = \{k_1, k_2, \dots, k_M\}$, the loss minimizes the negative log-likelihood of each index $k_m$ conditioned on its history $k_{<m}$ and the spatial context:$$\mathcal{L}_{feat} = \mathbb{E} \left[ - \sum_{m=1}^{M} \log_2 P_\psi(k_m \mid k_{<m}, \mathbf{h}_{spatial}, x) \right]$$where $P_\psi$ is the probability distribution predicted by the GRU model. For the geometry attributes, we adopt the bitrate loss formulation proposed in PCGS \cite{pcgs2025}, which employs trit-plane quantization for progressive encoding. Additionally, $\mathcal{L}_{scale}$ and $\mathcal{L}_{offset}$ serve as standard regularizers to prevent unbounded geometry growth.
$$\mathcal{L}_{rate} = \mathcal{L}_{feat} + \mathcal{L}_{hash} + \mathcal{L}_{scale} + \mathcal{L}_{offset}$$
\\
where $\mathcal{L}_{hash}$ is defined as in HAC++ \cite{hac++2025}.

\subsubsection{Quantization Loss}
The VQ-VAE parameters are updated by a separate, dedicated optimizer. The goal of this optimizer is to minimize the distance between the continuous feature and the quantized one. The RVQ-VAE objective $\mathcal{L}_{VQ}$ is the sum of a Feature Reconstruction Loss and a Codebook Commitment Loss:
$$\mathcal{L}_{rec} = \mathcal{L}_1(f_{cont}, f_{q})$$
$$\mathcal{L}_{commit} = \beta \| \mathbf{z}_e(\mathbf{x}) - \text{sg}(\mathbf{e}) \|_2^2$$
$$\mathcal{L}_{VQ} = \mathcal{L}_{rec} +\lambda_{commit} \mathcal{L}_{commit}$$

\subsection{Curriculum Learning}
Training Vector Quantized networks can be  unstable due to their non-differentiable nature, and a cold start with hard quantization often leads to codebook collapse and sub-optimal rendering quality \cite{zhao2024representation}, as it's significantly harder to converge. To mitigate this and ensure high-fidelity reconstruction, we implement a multi-stage curriculum learning strategy that gradually transitions the network from continuous to discrete representations.

\subsubsection{Phase 1: Continuous Feature Warm-up}
In the initial training phase (Steps $0$ to $T_{start}=10\text{k}$), we disable quantization entirely. The network optimizes the anchor features $\mathbf{f}_{cont}$ directly. To prepare the features for the distribution shift, we add small uniform noise to the scaling and offset parameters to simulate quantization error and improve decoder robustness.

\subsubsection{Phase 2: Soft Quantization Injection}

Between steps $T_{start}$ and $T_{end}=30\text{k}$, we linearly interpolate between the continuous features and their quantized counterparts. Let $\mathbf{f}_{q}$ be the output of the VQ-VAE. The feature used for rendering, $\mathbf{f}_{render}$, is computed as:$$\mathbf{f}_{render} = (1 - \beta) \cdot \mathbf{f}_{cont} + \beta \cdot \text{sg}[\mathbf{f}_{q} + (\mathbf{f}_{cont} - \text{sg}[\mathbf{f}_{cont}])]$$where $\beta$ is a time-dependent warmup factor that linearly increases from $0$ to $1$. This transition allows the set of $\Phi$ MLPs to progressively adapt to an increasingly quantized signal.

\subsubsection{Phase 3: Hard Quantization and Entropy Minimization}
After $T_{end}$, the network switches to Hard Quantization ($\beta=1$). On top of that, we enable the entropy model and add the rate loss $\mathcal{L}_{rate}$ to the objective.

\subsubsection{Codebook collapse mitigation}

To further mitigate codebook collapse and maximize representational capacity, we supplement our curriculum learning with an Exponential Moving Average stale code replacement strategy. Following the approach in SoundStream \cite{zeghidour2021soundstream}, if a code's usage falls below a predefined threshold, it is identified as "stale" and re-initialized by replacing it with a random vector from the current input batch. Additionally, we apply Layer Normalization at the output of the first projector to stabilize the feature distribution, ensuring that the latent vectors are appropriately scaled and centered before undergoing the quantization process, thereby improving codebook utilization.

\subsection{Progressive Transmission}
\subsubsection{Header and Base Layer ($m=1$)}
The initial transmission block consists of:

\begin{itemize}
    \item Binarized Spatial Hash Grid: To minimize the memory footprint of the context model, we binarize the parameters of the multi-resolution spatial hash grid, as proposed by HAC \cite{hac2024} and HAC++ \cite{hac++2025}, which are based on the methodology proposed by BiRF \cite{NEURIPS2023_aebf6284}.
    \item MLP Decoder Compression: The weights of the lightweight decoding MLPs ($\Phi$) are compressed using Zstandard \cite{rfc8878}.
    \item Base Visibility Mask: We explicitly encode the binary visibility state of each anchor and its associated Gaussian primitives for the base level, determining which primitives contribute to the coarse rendering.
    \item Anchors and Base Features: We encode the sparse anchor positions using Geometry Point Cloud Compression \cite{gpcc}. Alongside the geometry, we transmit the first quantization index $k_1$ for each active anchor, which will be decoded into the coarse latent feature on decoding. $$  \mathbf{f}^{1}_i = \text{Decoder}(k_1)$$
\end{itemize}

\subsubsection{Refinement Layers ($m > 1$)}
Subsequent data chunks transmit both feature residuals and geometry updates.

\begin{itemize}
    \item Incremental Visibility: Rather than re-transmitting the full visibility mask at every level, we employ Differential Mask Encoding. We compute the difference between the binary mask at level $m$ and level $m-1$, transmitting only the indices of newly activated Gaussians. This ensures zero redundancy for primitives that were already visible.
    \item Feature Refinement: For active anchors, the bitstream provides the residual indices $k_m$. The client then updates the latent features:$$  \mathbf{f}^{(m)}_i = \text{Decoder}(\sum_{j=1}^{m} k_j)$$
\end{itemize}

At time $t_{0}$, the client renders the base layer; at subsequent stages ($t_{1}$, $t_{2}$, ...), the client incorporates residuals and differential visibility masks, progressively improving fidelity without reloading the full scene.

\section{Experiments and Results}

\subsection{Experimental Setup}

\subsubsection{Datasets}
We evaluate SCAR-GS on synthetic and real-world benchmarks for neural rendering and Gaussian Splatting. For real-world performance evaluation, we use \textbf{MipNeRF360}~\cite{mipnerf360} to feature large, unbounded scenes suited for progressive compression evaluation. \textbf{BungeeNeRF}~\cite{bungeenerf}, provides additional unbounded outdoor scenes (Amsterdam, Bilbao, Hollywood, Pompidou, Quebec) extracted from Google Earth \cite{googleearth2026}.

\subsubsection{Baselines}
We compare SCAR-GS against state-of-the-art progressive compression methods for 3DGS: PCGS~\cite{pcgs2025} and GoDe~\cite{Sario2025GoDeGO}.

\begin{itemize}
    \item \textbf{PCGS \cite{pcgs2025}} achieves progressivity via \textit{trit-plane scalar quantization} with incremental mask transmission. It progressively increases both the number of Gaussian primitives and the precision of their scalar attributes at each Level of Detail (LOD).

    \item \textbf{GoDe \cite{Sario2025GoDeGO}} organizes Gaussians into hierarchical layers based on visibility heuristics. Progressivity is achieved through \textit{layer-wise primitive replication}, where each LOD adds more primitives rather than refining their attributes.
\end{itemize}

\begin{figure}
    \centering
    \includegraphics[width=0.75\linewidth]{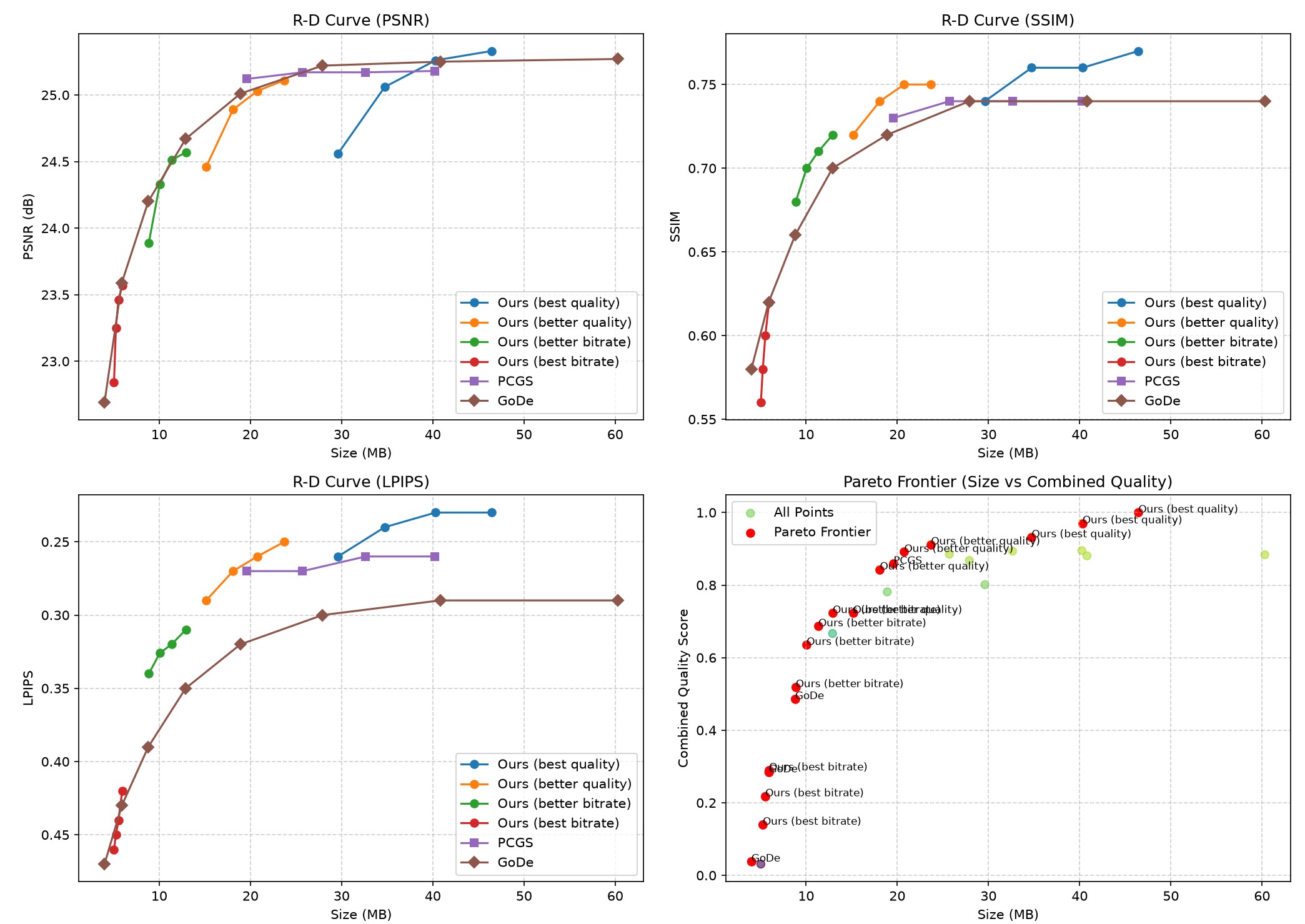}
    \caption{R-D curve of our method over different $\lambda_{ssim}$ (0.1, ..., 0.4) values in the Bicycle scene from the MipNeRF360 dataset. PCGS reaches slightly higher absolute PSNR/SSIM at small file sizes, where its low per-layer overhead dominates; SCAR-GS overtakes on rate of quality gain per added byte past 15 MB, and its Pareto frontier (bottom-right) stays populated across a wider size range, indicating more usable intermediate operating points for a streaming client choosing a bitrate mid-transmission.}
    \label{fig:placeholder}
\end{figure}

\begin{table}[h!]
\centering
\caption{Per-scene averaged results across the evaluated datasets after 40k training iterations.}
\resizebox{0.8\columnwidth}{!}{
\begin{tabular}{l c c c c c c c c c}
\hline
\textbf{Dataset} & \textbf{Method} & \textbf{Size (MB)} $\downarrow$ & \textbf{PSNR} $\uparrow$ & \textbf{SSIM} $\uparrow$ & \textbf{LPIPS} $\downarrow$\\
\hline
\multirow{12}{*}{Mip-NeRF360}
& Ours (ss0) & 11.71 & 25.20 & 0.7679 & 0.2812 \\
& Ours (ss1) & 13.54 & 26.54 & 0.7807 & 0.2565 \\
& Ours (ss2) & \textbf{15.36} & 27.09 & 0.8013 & 0.2463 \\
& Ours (ss3) & \textbf{19.73} & 27.27 & 0.8052 & 0.2416 \\
& PCGS (ss0) & 12.19 & \textbf{27.81} & \textbf{0.8079} & \textbf{0.2404} \\
& PCGS (ss1) & 15.4 & \textbf{27.93} & \textbf{0.8103} & \textbf{0.2377} \\
& PCGS (ss2) & 20.56 & \textbf{27.95} & \textbf{0.8109} & \textbf{0.2366} \\
& PCGS (ss3) & 25.34 & \textbf{27.96} & \textbf{0.8111} & \textbf{0.2361} \\
& GoDE (LOD 0) & \textbf{3.76} & 25.28 & 0.7189 & 0.4009 \\
& GoDE (LOD 2) & \textbf{8.63} & 26.63 & 0.7756 & 0.3407 \\
& GoDE (LOD 4) & 20.63 & 27.30 & 0.8040 & 0.2927 \\
& GoDE (LOD 7) & 73.73 & 27.42 & 0.8110 & 0.2684 \\
\hline
\multirow{6}{*}{BungeeNeRF}
& Ours (ss0) & \textbf{11.76} & 23.76 & 0.7832 & 0.2890 \\
& Ours (ss1) & \textbf{14.02} & 25.43 & 0.8300 & 0.2461 \\
& Ours (ss2) & \textbf{16.40} & 26.09 & 0.8460 & 0.2297 \\
& PCGS (ss0) & 12.97 & \textbf{26.89} & \textbf{0.8636} & \textbf{0.2234} \\
& PCGS (ss1) & 17.47 & \textbf{27.10} & \textbf{0.8696} & \textbf{0.2167} \\
& PCGS (ss2) & 22.82 & \textbf{27.15} & \textbf{0.8717} & \textbf{0.2132} \\
\hline
\end{tabular}
}
\label{tab:results}
\end{table}

Table ~\ref{tab:results} demonstrates that SCAR-GS provides consistent and monotonic improvements in perceptual quality as refinement layers are added. Compared to GoDe and PCGS, SCAR-GS achieves smoother quality gains with substantially lower storage growth, highlighting the advantage of feature refinement over primitive-based LOD strategies. We can also appreciate that SCAR-GS maintains consistent perceptual improvements across extremely large-scale outdoor scenes. While PCGS achieves higher absolute reconstruction quality at several operating points, \textbf{SCAR-GS provides smoother progressive refinement with moderate storage growth}, offering complementary operating characteristics for adaptive streaming.

\subsubsection{Implementation Details}

SCAR-GS is trained for 40k iterations. The RVQ-VAE codebooks are trained \emph{per scene}, jointly with the Gaussian Splatting optimization, using the training split of the corresponding dataset, without any external data or pretraining. The RVQ-VAE uses $N=4$ quantization stages with a dual-codebook design consisting of a base codebook and a shared residual codebook, each containing 1024 entries, as $log_2(1024)=10$. All experiments are conducted on NVIDIA A100 and H100 GPUs.

Entropy decoding with the GRU and spatial-query attention is performed once per transmission step, scales linearly with the number of active anchors and refinement layers, and is amortized over subsequent rendering, leaving runtime FPS unaffected. Results indicate that \textbf{SCAR-GS provides more compact representations for medium-to-large scenes} ( $> 10 MB $), while scalar quantization methods such as PCGS \cite{pcgs2025} perform better on smaller scenes. Encoding and decoding times are higher, but runtime efficiency was not the focus; further decoding optimization remains future work.


\section{Ablation Studies}
To validate the effectiveness of our architectural choices, we conducted ablation studies with $\lambda_{ssim} = 0.2$, isolating specific components to evaluate their individual contributions to compression efficiency and reconstruction quality. Due to time constraints, ablations are conducted on Bicycle, a representative large MipNeRF360~\cite{mipnerf360} scene; full multi-scene ablations are left for extended work.

\subsection{Architecture of the Entropy Model}

We evaluate the impact of the context model architecture on compression efficiency by comparing our proposed GRU with Spatial-Query Attention against three baselines: a standard MLP, a Branched MLP (separate heads for spatial and feature context), and a vanilla GRU without the spatial attention mechanism.

Unlike prior entropy models that apply generic attention, our spatial-query attention conditions residual history asymmetrically, using spatial embeddings as queries over decoded residual sequences to enable geometry-aware sequential probability estimation.

\begin{table}[h]
\centering
\caption{Ablation study on the architecture of the entropy model. Our proposed GRU with Spatial-Query Attention achieves the best compression rate (smallest size) and reconstruction quality. Spatial-query attention is the key component responsible for reducing entropy, not just improving reconstruction.}
\label{tab:entropy_ablation}
\begin{tabular}{lcccc}
\toprule
\textbf{Architecture} & \textbf{Size} $\downarrow$ & \textbf{SSIM} $\uparrow$ & \textbf{LPIPS} $\downarrow$ & \textbf{PSNR} $\uparrow$ \\
\midrule
MLP & 19.3 & 0.71 & 0.32 & 24.5 \\
Branched MLP & 22.4 & 0.72 & 0.31 & 24.6 \\
GRU & 21.9 & 0.71 & 0.30 & 24.4 \\
\textbf{GRU + Attn.} & \textbf{18.0} & \textbf{0.73} & \textbf{0.29} & \textbf{24.7} \\
\bottomrule
\end{tabular}
\end{table}

As shown in Table~\ref{tab:entropy_ablation}, our proposed architecture significantly outperforms all baselines. Simple MLPs cannot model sequential dependencies between residual codes, treating each quantization stage independently. The Branched MLP improves slightly by processing spatial and feature contexts separately, but fails to effectively fuse these modalities: the separate heads optimize independently without capturing their interaction.

The vanilla GRU successfully models temporal structure across residual layers but lacks spatial conditioning, achieving 21.9 MB at 0.71 SSIM. Without geometric context, probability predictions cannot adapt to local scene characteristics. Our Spatial-Query Attention mechanism bridges this gap by treating spatial embeddings as queries attending over GRU hidden states, allowing the network to dynamically weight residual history based on local geometry. This achieves 18.0 MB at 0.73 SSIM: a 7\% size reduction and 0.02 SSIM improvement over vanilla GRU, demonstrating that spatially-conditioned autoregressive modeling is essential for efficient entropy coding.

\subsection{Residual vs. Standard Vector Quantization}
A key design choice in SCAR-GS is using RVQ (progressive) over VQ (single-rate). We compared our RVQ approach against single-stage VQ with a larger 4096-entry codebook to match capacity.

\begin{table}[h]
\centering
\caption{Comparison between standard single-rate VQ and our progressive RVQ. RVQ yields superior rate-distortion performance. This confirms that progressiveness is not merely a transmission wrapper; RVQ itself improves compression over single-stage VQ.}
\label{tab:vq_vs_rvq}
\begin{tabular}{lcccc}
\toprule
\textbf{Architecture} & \textbf{Size} $\downarrow$ & \textbf{SSIM} $\uparrow$ & \textbf{LPIPS} $\downarrow$ & \textbf{PSNR} $\uparrow$ \\
\midrule
VQ & 20.8 & 0.72 & 0.29 & 24.4 \\
\textbf{RVQ} & \textbf{18.0} & \textbf{0.73} & \textbf{0.29} & \textbf{24.7} \\
\bottomrule
\end{tabular}
\end{table}

Table~\ref{tab:vq_vs_rvq} confirms that RVQ is superior for both streaming capability and compression efficiency. By decomposing the feature space into "coarse" base signals and "fine" residuals, RVQ enables the entropy model to learn more distinct, lower-entropy distributions for each stage. The base layer captures high-variance global structure with a broad probability distribution, while residual layers model progressively lower-entropy refinements with peaked distributions.

Standard VQ requires 20.8 MB to achieve 0.72 SSIM: 15\% larger than our RVQ at better quality (0.73 SSIM). This reflects the difficulty of optimizing entropy for large single-stage codebooks where the model must capture all feature complexity in one distribution without sequential context. Our autoregressive conditioning on previous quantizations produces inherently more compressible probability distributions.

\subsection{Gradient Propagation: Rotation Trick vs. STE}
We analyzed the impact of the gradient estimator used for the non-differentiable quantization step. We compared the STE \cite{bengio2013estimating} against the Rotation Trick \cite{fifty2025rotationtrick} implemented in our pipeline.

\begin{table}[h]
\centering
\caption{Impact of the gradient estimator on training stability and final quality. The Rotation Trick allows for better geometric capture on the gradient flow, leading to better reconstruction fidelity compared to the STE.}
\label{tab:grad_estimator}
\begin{tabular}{lcccc}
\toprule
\textbf{Estimator} & \textbf{Size} $\downarrow$ & \textbf{SSIM} $\uparrow$ & \textbf{LPIPS} $\downarrow$ & \textbf{PSNR} $\uparrow$ \\
\midrule
STE & \textbf{16.9} & 0.73 & 0.29 & 24.6 \\
\textbf{Rotation Trick} & 18.0 & \textbf{0.73} & \textbf{0.29} & \textbf{24.7} \\
\bottomrule
\end{tabular}
\end{table}

Table~\ref{tab:grad_estimator} shows that while STE produces 6\% smaller files, the Rotation Trick achieves marginally better PSNR (24.7 vs. 24.6) with identical perceptual metrics. More importantly, we observed more stable training dynamics with the Rotation Trick during our experiments.
The Rotation Trick injects geometric information about quantization error magnitude and direction into gradients by modeling the encoder-to-codebook relationship as a smooth linear transformation. This leads to more balanced codebook utilization and avoids local minima where certain entries dominate. The 6\% size increase likely reflects more conservative probability estimation when gradients carry geometric information, but improved training stability and consistent convergence justify this tradeoff for robust deployment across diverse scenes. Across ablation studies, we observe that architectural choices improving stability and representational robustness may incur modest increases in bitrate. These increases reflect tighter entropy modelling and improved generalisation rather than reduced compression effectiveness, and consistently result in superior perceptual quality and convergence behaviour.

\subsection{Impact of Curriculum Learning}
As mentioned in the methodology section, training RVQ-VAEs with hard quantization from initialization can be unstable. We evaluate our three-phase curriculum learning strategy against a cold-start approach.

\begin{table}[h]
\centering
\caption{Evaluation of the curriculum learning strategy. A "cold start" without warm-up leads to significant quality degradation, while our curriculum schedule ensures robust convergence.}
\label{tab:curriculum}
\resizebox{0.8\columnwidth}{!}{%
\begin{tabular}{lcccc}
\toprule
\textbf{Training Strategy} & \textbf{Size} $\downarrow$ & \textbf{SSIM} $\uparrow$ & \textbf{LPIPS} $\downarrow$ & \textbf{PSNR} $\uparrow$ \\
\midrule
Cold Start & \textbf{13.7} & 0.70 & 0.34 & 24.4 \\
\textbf{Curriculum Learning} & 18.0 & \textbf{0.73} & \textbf{0.29} & \textbf{24.7} \\
\bottomrule
\end{tabular}%
}
\end{table}

Table~\ref{tab:curriculum} demonstrates the disadvantage of cold-start training. Without gradual adaptation, the network prematurely commits to suboptimal codebook entries, causing codebook collapse where only a small subset of entries are actively used. The feature encoder learns to map all inputs to this limited subset, destroying representational capacity. Additionally, the scene decoder receives discrete inputs from initialization without the opportunity to learn smooth interpolation between codebook vectors.

Our curriculum learning (continuous warm-up (0-10k), soft quantization injection (10k-30k), and hard quantization refinement (30k-40k)) achieves 18.0 MB at 0.73 SSIM and 0.29 LPIPS. The 31\% storage increase versus cold start is necessary to avoid catastrophic quality loss: 0.03 SSIM improvement and 17\% LPIPS reduction. This validates that stable VQ training requires (1) warm initialization with continuous features, (2) gradual introduction of quantization constraints, and (3) progressive commitment to discrete representations.

\subsection{Spatial Context Representation}

We validated the design of our spatial hash grid. We compared a pure 3D Hash Grid against the Hybrid 2D+3D Grid, proposed by HAC++ \cite{hac++2025}.

\begin{table}[h]
\centering
\caption{Effectiveness of the spatial hash grid representation. The hybrid 2D+3D grid captures anisotropic correlations better than a pure 3D grid.}
\label{tab:spatial_grid}
\begin{tabular}{lcccc}
\toprule
\textbf{Spatial Grid} & \textbf{Size} $\downarrow$ & \textbf{SSIM} $\uparrow$ & \textbf{LPIPS} $\downarrow$ & \textbf{PSNR} $\uparrow$ \\
\midrule
3D Grid & \textbf{14.9} & 0.70 & 0.33 & 24.5 \\
\textbf{Hybrid Grid} & 18.0 & \textbf{0.73} & \textbf{0.29} & \textbf{24.7} \\
\bottomrule
\end{tabular}
\end{table}

Table~\ref{tab:spatial_grid} shows that the hybrid grid achieves 18.0 MB at 0.73 SSIM versus the pure 3D grid's 14.9 MB at 0.70 SSIM. The 21\% size increase is justified by substantial quality improvements: 0.03 SSIM gain and 12\% LPIPS reduction.

The hybrid design captures anisotropic spatial correlations: directional dependencies in scene structure. Many real-world scenes exhibit ground-plane dominated structure where lateral context (neighboring buildings, terrain features) differs fundamentally from vertical context (sky, height variations). Pure 3D grids treat all directions equally, failing to model these directional patterns.

When predicting residual indices, the hybrid grid allows the entropy model to distinguish high-correlation directions (lateral neighbors) from low-correlation directions (vertical), achieving tighter probability distributions, translating to more accurate probability estimation despite the additional hash grid parameters, improving both compression efficiency and reconstruction fidelity.

\section{Conclusion}

Although SCAR-GS does not model dynamic scene geometry, it addresses a practical temporal problem in spatial multimedia: how scene quality evolves during progressive delivery. Future work will explore a network streaming framework for SCAR-GS, including view-dependent adaptive transmission under varying bandwidth conditions. Additionally, exploring improved RVQ-VAE training objectives that more tightly preserve original feature structure may further enhance reconstruction fidelity and perceptual quality.

\subsubsection*{Acknowledgements.}
This work was supported by the NUS SRI Programme and the Department of Computer Science, NUS School of Computing.

%
%
\bibliographystyle{splncs04}
\bibliography{main}
\end{document}